\documentclass{article}

\usepackage{arxiv}
\usepackage{booktabs} 
\usepackage{amssymb}
\usepackage{amsmath}
\usepackage{tabularx}
 \usepackage{graphicx}
 \usepackage{longtable}
  \usepackage{multirow}
  \usepackage{hyperref}
 \usepackage[normalem]{ulem}
 \useunder{\uline}{\ul}{}

\newcommand{\mt}[1]{\mbox{\texttt{#1}}}

\title{
SQuAP-Ont: an Ontology of Software Quality Relational Factors from Financial Systems  
}

\author{
Paolo Ciancarini\\
University of Bologna, Italy and Innopolis University, Russia\\
\texttt{paolo.ciancarini@unibo.it}
\And
Andrea Giovanni Nuzzolese\\
STLab, Institute of Cognitive Sciences and Technologies, National Research Council, Rome, Italy\\
\texttt{andreagiovanni.nuzzolese@cnr.it}
\And
Valentina Presutti\\
STLab, Institute of Cognitive Sciences and Technologies, National Research Council, Rome, Italy\\
\texttt{valentina.presutti@cnr.it}
\And
Daniel Russo\\
Lero - The Irish Software Research Center \& School of Computer Science \& Information Technology \\
University College Cork, Cork, Ireland\\
\texttt{daniel.russo@lero.ie}
}

\begin{document}
\maketitle

\begin{abstract}
Quality, architecture, and process are considered the keystones of software engineering. 
ISO defines them in three separate standards. However, their interaction has been scarcely studied, so far. 
The SQuAP model (Software Quality, Architecture, Process) describes twenty-eight main factors that impact on software quality in banking systems, and each factor is described as a relation among some characteristics from the three ISO standards. 
Hence, SQuAP makes such relations emerge rigorously, although informally. 
In this paper, we present SQuAP-Ont, an OWL ontology designed by following a well-established methodology based on the re-use of Ontology Design Patterns (i.e. ODPs). SQuAP-Ont formalises the relations emerging from SQuAP to represent and reason via Linked Data about software engineering in a three-dimensional model consisting of quality, architecture, and process ISO characteristics. 
\end{abstract}
 
\section{A three-dimensional view on software quality}
\label{sec:intro}
Industrial standards are widely used in the software engineering practice: they are built on pre-existing literature and provide a common ground to scholars and practitioners to analyze, develop, and assess software systems. As far as software quality is concerned, the reference standard is the ISO/IEC 25010:2011 (ISO quality from now on), which defines the quality of software products and their usage (i.e., in-use quality). The ISO quality standard introduces eight \emph{characteristics} that qualify a software product, and five characteristics that assess its quality in use. A characteristic is a parameter for measuring the quality of a software system-related aspect, e.g., reliability, usability, performance efficiency. The quantitative value associated with a characteristic is measured employing metrics that are dependent on the context of a specific software project and defined following the established literature.

The ISO quality standard only focuses on the resulting software product without explicitly accounting for the \emph{process} that was followed or the implemented \emph{architecture}. However, there is wide agreement~\cite{pressman2014software} about the importance of the impact of three combined dimensions: software quality, software development process, and software architecture, on the successful management and evolution of information systems. In this respect, the industrial standard ISO/IEC 12207:2008 defines a structure for the software process life cycle, and outlines the tasks required for developing and maintaining software \cite{singh1996international}. Regardless of the chosen methodology (i.e., Agile or Waterfall ones \cite{pressman2014software}), this standard identifies the relevant concepts of the life cycle and provides a useful tool for software developers to assess if they have undertaken all recommended actions or not. Each lifecycle concept can be evaluated according to its maturity level through established metrics, e.g., the Capability Maturity Model Integration (CMMI)~\cite{team2002capability}. As for the architectural dimension, the ISO/IEC 42010:2011 standard provides a glossary for the relevant objects of software architecture. 
Concerning software architecture evaluation, intended as a way to achieve quality attributes (i.e., maintainability and reliability in a system), 
some approaches have emerged, the most prominent being ATAM, proposed by the Software Engineering Institute \cite{kazman1994saam,CKK02,bengtsson2004architecture,bellomo2015toward}.
Typical research in this domain is about how architectural patterns and guidelines impact software components and configurations \cite{garlan1995introduction}. A survey study \cite{dobrica2002survey} analyzes architectural patterns to identify potential risks, and to verify the quality requirements that have been addressed in the architectural design of a system.

The mutual relations among the three dimensions and their impact on the quality of software systems have been barely addressed in the literature, but a recent empirical study~\cite{Russo2017metaquality} in the domain of Software Banking Systems pointed out the importance of those relations. The study involved 13 top managers of the IT Banking sector in the first phase and 124 additional domain experts in a second validation phase. The result of such a study was a model named SQuAP that describes these relations in terms of \emph{quality factors}. 
According to \cite{Gorla2010b}, the information available to guide and support the management of software quality efforts is a critical success factor for IT domains. Considering the broad coverage of its empirical provenance, a formalization of SQuAP may serve as a reference resource and practical tool for scholars and practitioners of software engineering, to assess the quality of a software system, to drive its development to meet a certain quality level, as well as to teach software engineering. 

The SQuAP model builds on the concept of \emph{quality factor}: a $n$-ary relation between software quality characteristics that cover the three dimensions of software product, process, and architecture, based on the three reference standards ISO/IEC 25010:2011, ISO/IEC 12207:2008, and ISO/IEC 42010:2011 respectively.
A SQuAP quality factor can be described as a complex quality characteristic (or \emph{parameter}) that provides a three-dimensional view for assessing software quality. The model identifies twenty-eight quality factors.

Our contribution consists in a resource named SQuAP-Ont, an ontology that formally represents the concept of quality factor by reusing existing ontology design patterns (e.g., Description and Situation~\cite{DBLP:books/ios/p/PresuttiG16,DBLP:journals/aamas/Gangemi08}), instantiates all factors identified so far, and axiomatizes them in order to infer measurable factors based on the characteristics available at hand. Besides, the ontology has been annotated with OPLa\footnote{http://ontologydesignpatterns.org/opla/} (Ontology Design Pattern representation language) to increase its reusability. SQuAP-Ont is publicly available online\footnote{ https://w3id.org/squap/} with accompanying documentation that describes the factors, under a CC-BY-4.0 license. 

In the rest of the paper, after discussing relevant related work (Section \ref{sec:state}), we provide additional details about the SQuAP model by presenting two sample factors in Section \ref{sec:model}. We describe the SQuAP-Ont ontology: its main concepts and axioms, the adopted design methodology and the reused ontology design patterns in Section \ref{sec:ont}. In Section \ref{sec:example} we provide examples of how to use it; and discuss the resource potential impact (Section \ref{sec:impact}) before concluding and identifying future developments, in Section \ref{sec:conclusions}.

\section{Related work}
\label{sec:state}
The use of ontologies in the software engineering domain is very common~\cite{calero2006ontologies,zhao2009ontology,kabaale2017axiom}. 
The ISO standards referenced in Section \ref{sec:intro} have been the subject of several ontological studies. For example, useful guidelines for their ontological representation are proposed by Henderson \textit{et al.} (2014) and Gonzalez \textit{et al.} (2016) \cite{henderson2014ontology,gonzalez2016ontology}.

An ontology-based approach to express software processes at the conceptual level was proposed in Liao \textit{et al.} (2015) \cite{liao2005software} and implemented in e.g., Soydan \& Kokan (2006) \cite{soydan2006owl} for CMMI \cite{chrissis2003cmmi}. Software quality attributes have been modeled in Kayed \textit{et al.} (2009) \cite{kayed2009towards}, while an ontology for representing software evaluation is proposed in Castro \textit{et al.} (2010)
\cite{upm6915}. Finally, a formalisation of the ISO 42010, describing software architecture elements, is developed in Emery \& Hiliard (2009) \cite{emery2009every} and in Kumar \& Prabhakar (2010) \cite{kumar2010pattern}; and Antunes \textit{et al.} (2013) \cite{antunes2013using} argues that different architecture domains can be integrated and analyzed through the use of ontologies.

Most of the works mentioned above focus on a strict representation of standards in terms of ontologies~\cite{liao2005software,soydan2006owl,chrissis2003cmmi,emery2009every}. Other scholars~\cite{kayed2009towards,upm6915} provide only preliminary ontological solutions for modelling quality characteristics or software evaluation and, to the best of our knowledge, they overlook the reuse of ontology design patterns.
In contrast, our work focuses on the relation between the different ISO standards (system quality, software development process, and software architecture) for supporting the assessment of software system quality, with the added value of following a rigorous pattern-based ontology design approach.

Also at a higher level, an ontology to harmonize different hierarchical process levels through multiple models such as CMMI, ISO 90003, ITIL, SWEBOK, COBIT was presented in Pardo \textit{et al.} (2012) \cite{pardo2012ontology}.

Ontologies referred to software quality focus primarily on quality attributes \cite{kayed2009towards}.
One quality evaluation, based on the ISO 25010 standard, is enhanced by taking into consideration several object-oriented metrics \cite{motogna2015improving}.
Similarly, Castro \textit{et al.} (2010) reuse current quality standards and models to present an ontology for representing software evaluations \cite{upm6915}.

Similarly, scholars advanced an ontology for 
the ISO 42010 standard regarding software architecture  \cite{emery2009every}.
An ontological representation of architectural elements has also been expanded by Kumar \& Prabhakar (2010) \cite{kumar2010pattern}.
With particular reference to the architecture rationale, some visualization and comparison techniques with semantic web technologies have been proposed in literature \cite{lopez2009visualization}.
Moreover, scholars showed that different architecture domains could be integrated and analyzed through the use of ontologies \cite{antunes2013using}

Finally, the Semantic Web community proposed also guidelines regarding the representation of ISO standards of software engineering with ontologies \cite{henderson2014ontology,gonzalez2016ontology}.
These two papers use a domain ontology, proposing the creation of a single underpinning abstract domain ontology, from existing ISO/IEC standards.
According to the authors, the adoption of a single ontology will permit the re-engineering of existing International Standards as refinements from this domain ontology so that these variously focused standards may inter-operate.

\section{Relational quality factors: the SQuAP Model}
\label{sec:model}
The motivation for developing SQuAP is based on the understanding, in the software engineering and information systems communities, that assessing software quality for contemporary information systems requires to take into consideration the relations among different dimensional perspectives, namely: software quality, process, and architecture \cite{pressman2014software}.
Accordingly, we conducted an empirical study in the banking sector \cite{russo2018meta}.
The financial and banking industry is particularly useful to explore information systems quality issues for several reasons.
First, it is mission-critical, i.e., the failure of even one system would lead to unpredictable consequences.
Thus, the top manager showed increasing concerns about the quality and sustainability of their information systems \cite{Russo2017metaquality}.
Secondly, most institutions, such as those involved in the study, are multinational, sharing the same systems and the same concerns.
Finally, this sector is highly connected and regulated centrally by banking authorities.
So, many functionalities and requirements are well defined by such authorities, which means that the entire industry is using similar software products. 

Therefore, to develop the SQuAP quality meta-model, we executed our research according to the following steps.
In the first phase, based on the Delphi method~\cite{dalkey1969delphi}, we involved 13 top managers of this sector to express their most significant software quality concerns. The result was a set of distinct 28 quality factors emerging from the elicited concerns, after a consensus-based negotiation phase (part of the Delphi method). 
In a second phase, we involved 124 domain experts that validated the 28 factors with a high level of agreement.
Each factor has been then linked to several characteristics or elements defined in the three different standards, i.e., ISO 25010, ISO 42010, and ISO 12207, for software quality, architecture, and process respectively. 
We followed the theoretical coding approach by Strauss \& Corbin (1997) \cite{strauss1997grounded} to map the factors to the ISO standards\footnote{Standards are \textit{de facto} second-order theories, built on grounded pre-existing ones and shared among scholar's and practitioner's communities.}.

\begin{figure*}[ht!]
\centering
\includegraphics[scale=0.4]{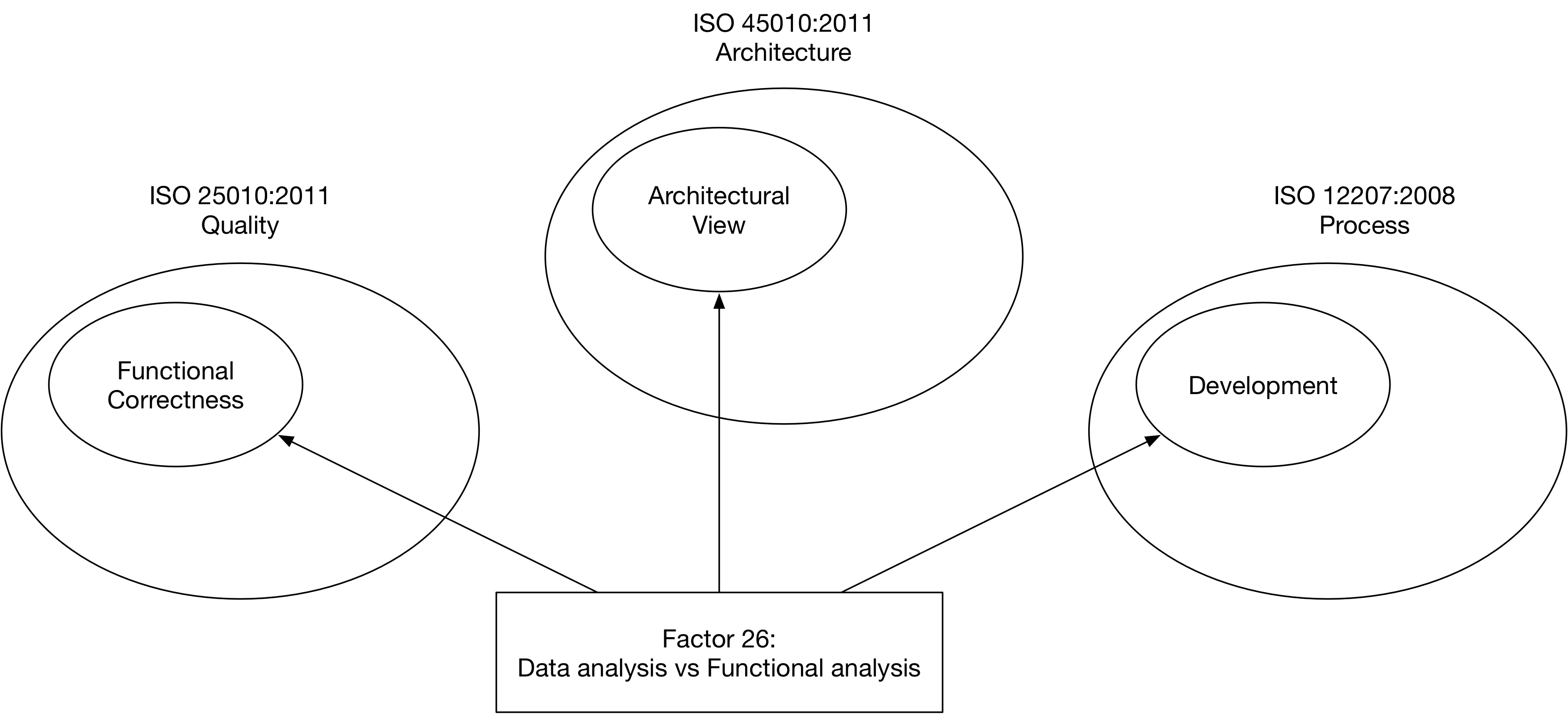}
\caption{Factor 26: Data analysis vs. Functional analysis. This factor is defined as a relation between three quality characteristics of a software project: Functional Correctness (ISO 25010), Architectural View (ISO 45010), and Development (ISO 12207).}
\label{fig:F26}
\end{figure*}

The selection criteria and demographics of the experts involved in both phases, the inclusion and exclusion criteria, the agreement figures as well as the industry coverage are explained in Russo \textit{et al.} (2018) \cite{russo2018meta}. In the same paper, we provide further details about the methodological approach and an exhaustive description of the 28 factors. A list of them are also published on the resource website\footnote{\url{https://w3id.org/squap/documentation/factors.html}} along with a short textual description, and by tables depicting their mappings to ISO standards. 

The 28 SQuAP factors have been rigorous, although informally defined in our previous work \cite{russo2018meta}. These definitions are the primary input for the development of SQuAP-Ont; hence, it is relevant to report here at least one them. The definition of a factor consists of a set of quotes from the experts involved in the study followed by an analysis (based on theoretical coding) on what are the main characteristics and elements from the standards that emerged as components of the factor.
We choose randomly one factor (26) to explain the underlying logic of the factor mapping. 

\textbf{Factor 26: Data analysis vs Functional analysis}.
This factor explores whenever poor data analysis influences functional analysis and so, system integrity.
\textit{``The ``functional centric'' view is quite misleading since it relegates the importance of data. Data give the ``static view'' of existing functionalities, which is very important for the functional analysis. One software product may have all possible functionalities required by the user but lacking fundamental data its deployment and system integration becomes impossible''}, said one surveyed expert.
Moreover, \textit{``data analysis skills are generally lacking and poorly used in functional analysis''}, affirmed another one.
This issue is present market-wide.
\textit{``In my opinion, in the market, there is a lacking perception about the importance of data analysis as the preliminary phase of functional analysis''}.
However, other experts disagree.
\textit{``Saying that poor analysis is due to poor data understanding is a quite generic (it is obvious that data processing is the main IT goal) and old issue''}. 
Other issues are also relevant to understand the factor.
\textit{``Also knowing what different data means is important''}.
Also, \textit{``it is not only an issue of poor technical skills''}.
Furthermore, \textit{``personally, I saw poorer knowledge of bank's operation processes''}.
There was a shift after 2010, which give fascinating insights.
\textit{``It is true for applications developed before 2010. Data governance is now more relevant, and data analysis is performed before the functional one. So I see a clear discontinuity with the past''}.

{\bf Theoretical coding analysis}: experts stressed the importance of data and functional analysis, impacting on the dimensions of {\bf Functional Correctness} (software quality), {\bf Development} (software process), and {\bf Architecture View} (software architecture). 
They refer to the functional suitability of applications through correctness.  
This impacts the development process, which is supported by such an analysis.
The architecture view addresses the concern of a suitable system held by the system's stakeholders. Figure \ref{fig:F26} shows a graphical representation of Factor 26.

\section{SQuAP-Ont: an OWL formalisation of SQuAP}
\label{sec:ont}
In this section we first provide details about the ontology design methodology adopted (cf. Section~\ref{sec:ont_design_tehodology}), then we describe the SQuAP Ontology (cf. Section~\ref{sec:ont_description}) and we provide its formalisation (cf. Section~\ref{sec:ont_formalisation}). Finally, we provide the implementation details of the ontology (cf. Section~\ref{sec:ont_implementation_details}).
\subsection{Design methodology}
\label{sec:ont_design_tehodology}

\begin{figure*}[ht!] \centering
\includegraphics[scale=0.47]{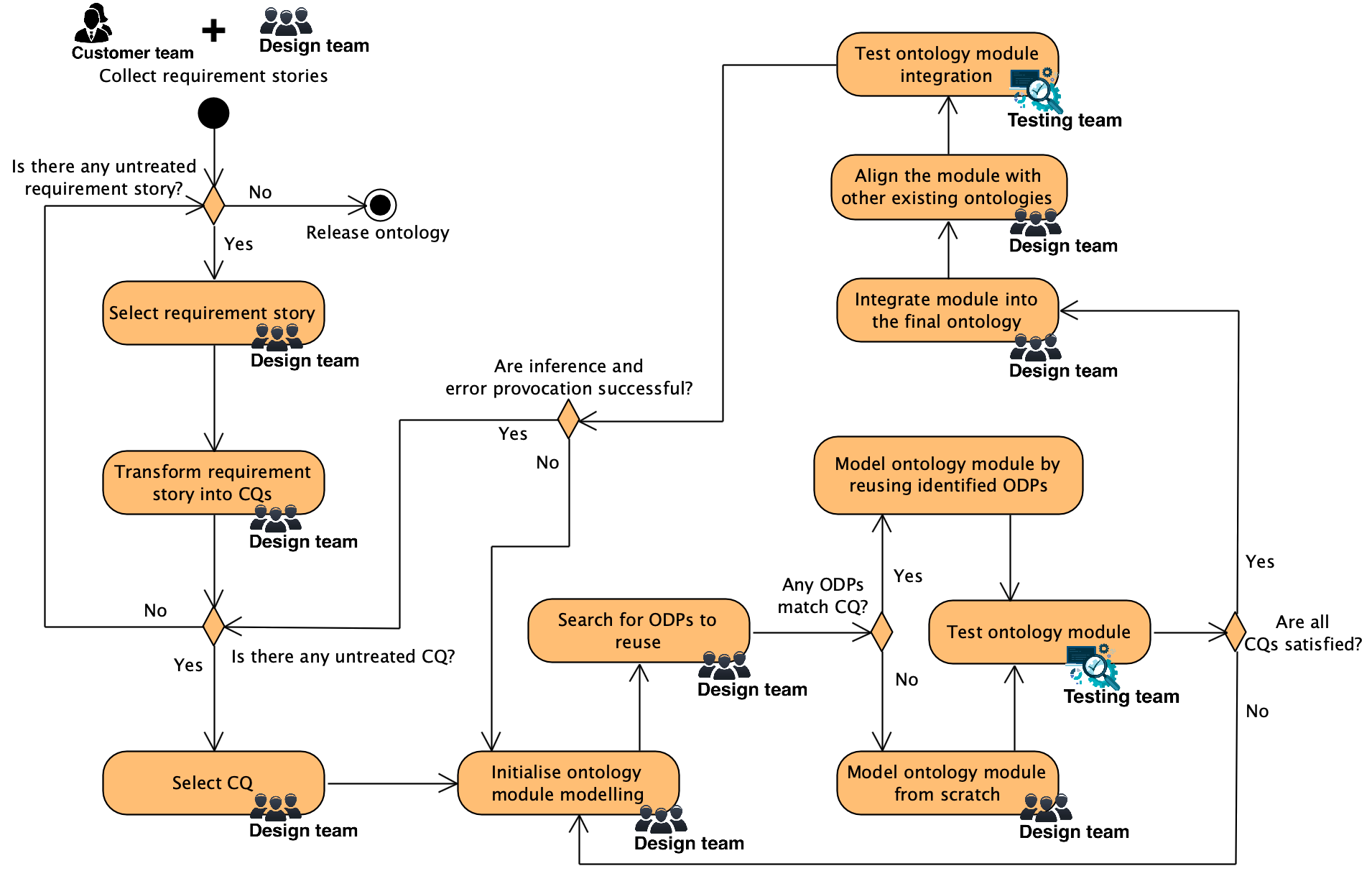}
    \caption{The XD methodology as implemented for modelling the SQuAP ontology.}
    \label{fig:xd-extension}
\end{figure*}

The SQuAP Ontology (SQuAP-Ont) is designed by reusing ontology design patterns (ODPs)~\cite{Gangemi2009} according to an extension of the eXtreme Design methodology~\cite{Blomqvist2010}. We opt for an ODP-based methodology as there are shreds of evidence~\cite{Blomqvist2010} that the reuse of ODPs (i) speeds up the ontology design process, (ii) eases design choices, (iii) produces more effective results in terms of ontology quality, and (iv) boosts interoperability. This extension has been first described in~\cite{Presutti2016} and mainly focuses on providing ontology engineers with clear strategies for ontology reuse. According to the guidelines provided by~\cite{Presutti2016}, we adopt the {\em indirect re-use}: ODPs are reused as templates. Hence, ODPs are specialized by specific ontology modules developed as components of the final ontologies instead of being directly reused and linked by those modules. For example, if we want to reuse the object property \texttt{region:hasParameter} from the Region ODP, then we define the object property \texttt{:hasParameter} into the target ontology and we formally annotate such indirect reuse employing the OPLa ontology~\cite{Hitzler2017}. Nevertheless, the ontology guarantees interoperability by keeping the appropriate alignments with the external ODPs and provides extensions that satisfy more specific requirements. 

The XD extension implemented in this work has been successfully adopted in several ontologies and linked open data projects so far. Examples are~\cite{Nuzzolese16,Peroni16,Lodi2017,Gangemi17,Carriero19}.  This extension implements an iterative and incremental approach to ontology design that involves three different actors: (i) a \emph{design team}, in charge of selecting and implementing suitable ODPs as well as to perform alignments and linking; (ii) a \emph{testing team},  disjoint from the design team, which takes care of testing the ontology; (iii) a \emph{customer team}, who elicits the requirements that are translated by the design team and testing team into ontological commitments (i.e., competency questions and other constraints) that guide the ontology development. Figure~\ref{fig:xd-extension} depicts the UML activity diagram that represents the XD extension described in Presutti \textit{et al.}~\cite{Presutti2016} as implemented in this work. The diagram is extended by explicitly identifying the actors that are associated with each action they are involved in. The first action of the methodology is the collection of requirements that involves both the customer and the design team. At this stage, the requirements are recorded as {\em stories}, which are typically used in agile software development for communicating requirements from the customer to the development team. After requirement stories are recorded, the design team starts to transform them into {\em competency questions}~\cite{Gruninger1995} (CQs). CQs represent the ontological commitments that drive the ontology development. Table \ref{table:cqs} reports the CQs, identified by analysing the SQuAP model (cf. Section \ref{sec:model}) and by discussing with domain experts (i.e., the customer team in our context). 

\begin{table}[h!]
\centering
\caption{Competency questions used for modelling SQuAP-Ont.} 
\label{table:cqs} 
\begin{tabular}{l | p{14cm} }
    \textbf{ID} & \textbf{Competency question}\\ \hline
    
\hline
 CQ1 & What are the quality characteristics of a software system at software, process, and architectural level?\\ \hline
 CQ2 & What are the factors, the assessment of which, is affected by a certain quality characteristic?\\ \hline
 CQ3 & What are the quality characteristics that affect the assessment of a certain factor?\\ \hline
 CQ4 & What is the unit of measure (i.e., metric) associated with a certain quality characteristic?\\ \hline
 CQ5 & What is the value computed for assessing a certain quality characteristic?\\ \hline
\end{tabular} 
\end{table}

The next action is about the design team looking for possible ODPs to reuse by analysing the resulting CQs. Those ODPs, if available, are reused for designing the modules of the ontology that addresses the specific CQs. When an ontology module is ready, the testing team performs the validation by assessing its fulfilment to the CQs. This validation is performed by (i) converting the CQs to SPARQL queries and (ii) executing those queries on a data sample, which is modelled according to the target ontology module. If the validation is successful, the design team integrates the ontology module in the final ontology.
Additionally, the design team provides alignments with related external ontologies and vocabularies in the Semantic Web for boosting interoperability. Then, the testing team performs a set of integration tests aimed at (i) validating the logical consistency of the ontology and (ii) its ability to detect errors by deliberately introducing contradictory facts. Both checks can be performed by using a DL reasoner, such as HermiT\footnote{\url{http://www.hermit-reasoner.com/}}, Pellet\footnote{\url{https://github.com/stardog-union/pellet}}, etc. We remark that the testing based on the (i) validation of the CQs, (ii) the check of the logical consistency, and (iii) the error provocation follows the methodology designed by Blomqvist \textit{et al.} (2012)  \cite{Blomqvist2012}. If the integration tests succeed, then the design team performs another iteration of the process by selecting an untreated CQ. If no untreated CQ is available, then the iteration consists of the selection of an untreated requirement story. Finally, after several iterations and when no untreated requirement story is available, the process ends. Accordingly, the ontology is released.

\subsection{Ontology description}
\label{sec:ont_description}

\begin{figure*}[ht!] \centering
\includegraphics[scale=0.42]{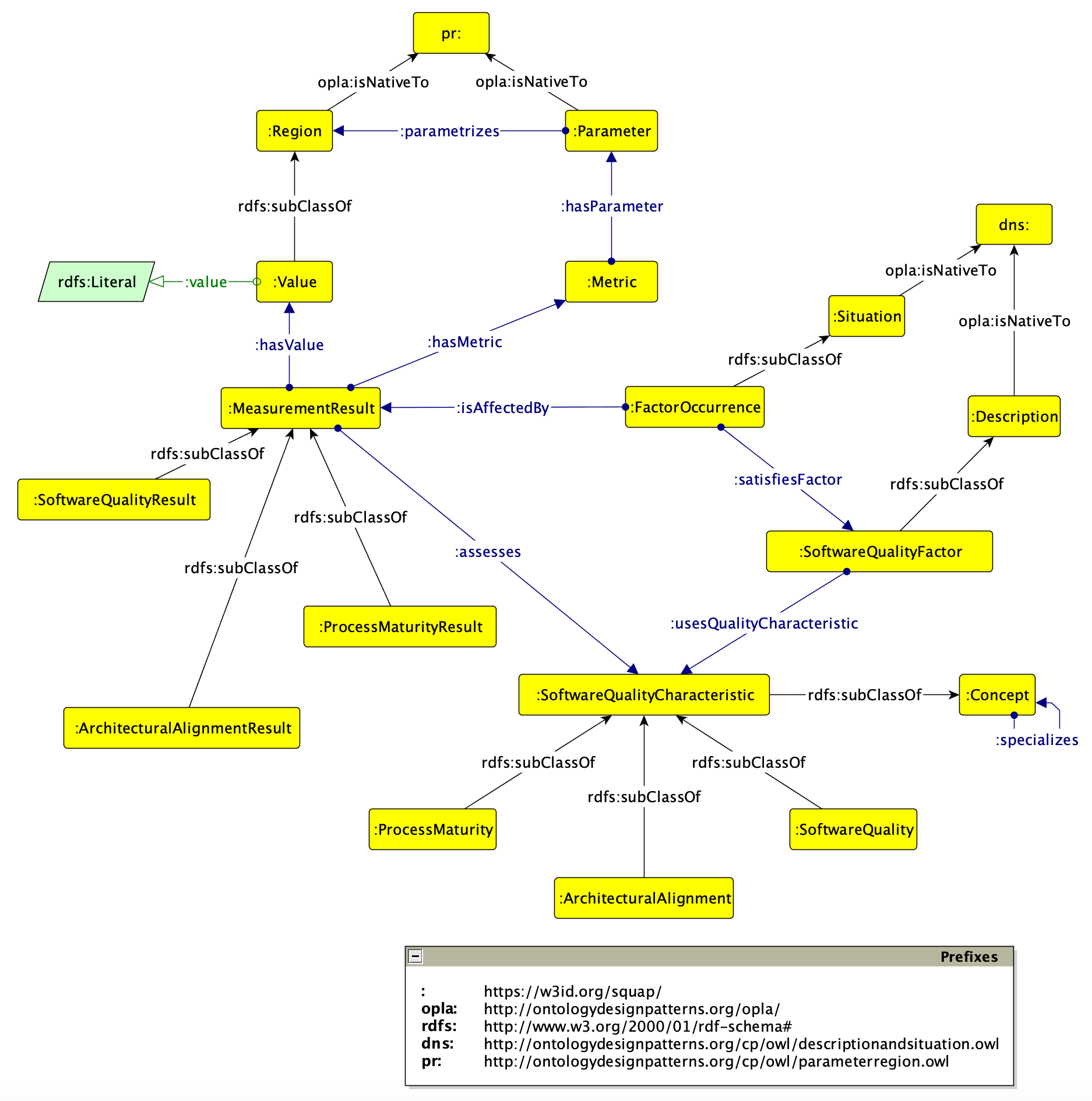}
    \caption{Core classes of SQuAP-Ont.}
    \label{fig:ontology}
\end{figure*}

Figure~\ref{fig:ontology} shows a diagram of SQuAP-Ont. We use the namespace \texttt{\url{https://w3id.org/squap/}}. 
SQuAP-Ont re-uses as templates the following ontology design patterns~\cite{DBLP:books/ios/p/PresuttiG16}: Description and Situation (D\&S)\footnote{\url{http://ontologydesignpatterns.org/cp/owl/descriptionandsituation.owl}}~\cite{Gangemi2003}, and Parameter Region\footnote{\url{http://ontologydesignpatterns.org/cp/owl/parameterregion.owl}}.

The D\&S pattern allows representing the conceptualisation of a $n$-ary relation (i.e., description) and its occurrences (i.e., situations) in the same domain of discourse. For example, it is used for representing the description of a plan (D) and its actual executions (S), the model of disease (D, e.g., its symptoms) and the actual occurrence of it in a patient (S), etc. SQuAP-Ont reuses this pattern for modelling quality factors with the class \texttt{:Software\-Quality\-Factor}, a subclass of \texttt{:Description}. The actual occurrences of quality factors assessed in a specific software project are modelled with the class \texttt{:Factor\-Oc\-cur\-rence}, a subclass of \texttt{:Situation}. Both \texttt{:Description} and \texttt{:Situation} are core elements of the D\&S pattern. According to the D\&S pattern a \texttt{:Description} defines a set of \texttt{:Concept}s. In the context of SQuAP-Ont we say that a \texttt{:Software\-Quality\-Factor} uses a set of \texttt{:Software\-Quality\-Char\-ac\-ter\-istic}. This relation is modelled by the property \texttt{:uses\-Quality\-Characteristic}. We model three types of \texttt{:Software\-Quality\-Char\-ac\-ter\-istic}: \texttt{:Software\-Quality}, \texttt{:Ar\-chi\-tec\-tural\-Align\-ment}, and \texttt{:Process\-Mat\-ur\-ity}. They classify the characteristics associated with the three different ISO standards and their own perspectives, i.e., software quality, architecture, and process. 
In a similar way, a set of entities are in the setting provided by a \texttt{:Situation}. In the context of SQuAP-Ont we say that a set of \texttt{:Meas\-ure\-mentRes\-ult} affects the assessment of a \texttt{:Factor\-Oc\-cur\-rence}. We model three types of \texttt{:Meas\-ure\-mentRes\-ult} with the classes 
\texttt{:Software\-Quality\-Meas\-ure\-ment\-Res\-ult}, \texttt{:Ar\-chi\-tec\-tural\-Align\-ment\-Res\-ult}, and \texttt{:Process\-Mat\-ur\-ity\-Res\-ult} which are instantiated with result measurements computed for assessing the quality characteristics of a specific software system. A \texttt{:Meas\-ure\-ment\-Res\-ult} has a \texttt{:Value} and a reference \texttt{:Metric}. For example, we may want to represent that the \textit{Reliability} of a software system is associated with a specific degree value according to a certain metric. This part of the model reuses the Parameter Region ontology design pattern as template. 

In D\&S each entity that is in the setting of a \texttt{:Situation} is classified by a \texttt{:Concept}. In the context of SQuAP-Ont we specialized this relation by saying that a \texttt{:Meas\-ure\-mentRes\-ult} assesses a \texttt{:Software\-Quality\-Char\-ac\-ter\-istic}. Based on the \texttt{:Meas\-ure\-mentRes\-ult}s that compose a \texttt{:Factor\-Oc\-cur\-rence} it may satisfy one or more \texttt{:Software\-Quality\-Factor}s (cf. modelled by the property \texttt{:satisfiesFactor}.

\texttt{:Software\-Quality\-Factor}s are represented in SQuAP both as individuals and classes, by exploiting OWL punning. Punning implements metaclassing in OWL and allows to interpret the same ontology entity either as a class or an instance of a metaclass depending on the syntactic context. This mechanism makes ontology modelling more similar to the way humans communicate knowledge by using natural language. Punning evoke verbal jokes (i.e., pun), which are typically used in natural language to emphasise a particular fact. 
When quality factors are interpreted as classes, then it is possible to introduce instances of those classes. Hence, it is possible to describe specific quality factors that occur for contextualising the quality of certain software. For example, it is possible to describe the quality resulting from the analysis of a specific software by introducing a specific individual of the factor \texttt{factor:QualityVsRequirements} represented as a class. On the contrary,
when quality factors are interpreted as individuals, then it is possible to treat them as instances of the metacass \texttt{:Software\-Quality\-Factor}. Hence, it is possible to predicate them. In the latter case, we can use the SQuAP-Ont to model a knowledge graph that provides facts about the quality factors from a general perspective. For example, we might state that 
\texttt{factor:Quality\-Vs\-Re\-quire\-ments} \texttt{dul:as\-sociated\-With} \texttt{factor:Qua\-li\-ty\-Vs\-Ti\-me\-And\-Budget}. Additionally, a clear benefit from modelling factors with punning is the possibility to use both DL axioms or rules (e.g., SPARQL CONSTRUCT) to infer new knowledge. All factors identified by the SQuAP model are instantiated in the ontology. SQuAP-Ont models the three types of \texttt{:Software\-Quality\-Char\-ac\-ter\-istic}s in a similar way: a set of individuals extracted from the SQuAP model, according to the three reference ISO standards (cf. Section~\ref{sec:model}), are included in the ontology. They are also modelled as classes to make the ontology extensible with possible specific axioms.
Furthermore, \texttt{:Software\-Quality\-Char\-ac\-ter\-istic}s are organized hierarchically through the object property \texttt{:specializes}, which is declared as transitive. The use of both \texttt{rdfs:subClassOf} and \texttt{:specializes} allows to represent hierarchical relations among \texttt{:Software\-Quality\-Char\-ac\-ter\-istic}s both when they are interpreted as classes and when they are interpreted as individuals. The latter is beneficial for defining and reasoning on software quality characteristics organized taxonomically within a controlled vocabulary, similarly to the taxonomic relations among concepts in SKOS (i.e., \texttt{skos:borrower}).

As aforementioned, SQuAP-Ont is annotated with the OPLa ontology~\cite{Hitzler2017} for explicitly indicating the reused patterns. We use the property \texttt{opla:reuses\-Pattern\-As\-Template} to link SQuAP-Ont to the two patterns we adopted as template, i.e., D\&S and Parameter Region. Similarly, we use the property \texttt{opla:isNativeTo} to indicate that certain classes and properties of SQuAP-Ont are core elements of specific ontology patterns. These annotations enable the automatic identification of the patterns reused by SQuAP-Ont, e.g., with SPARQL queries, hence facilitating the correct reuse of the ontology.

Finally, SQuAP is aligned, using an external file, to DOLCE+DnS UltraLight~\footnote{http://www.ontologydesignpatterns.org/ont/dul/DUL.owl}. Tables~\ref{table:alignments_class} and~\ref{table:alignments_props} report the alignments axioms between the classes and the properties of the two ontologies, respectively.

\begin{table}[h!]
\centering
\caption{Align\-ments between the classes of SQuAP-Ont and DOLCE UltraLight.} 
\label{table:alignments_class} 
\begin{tabular}{ l| c | >{\raggedleft\arraybackslash}p{3cm} }
    \textbf{SQuAP class} & \textbf{Align. axiom} & \textbf{DOLCE class}\\ \hline
    
\hline
  \texttt{:Region} & \texttt{owl:equivalentClass} & \texttt{dul:Region} \\ \hline
  \texttt{:Value} & \texttt{owl:subClassOf} & \texttt{dul:Amount} \\ \hline
  \texttt{:Parameter} & \texttt{owl:equivalentClass} & \texttt{dul:Parameter} \\ \hline
  \texttt{:Concept} & \texttt{owl:equivalentClass} & \texttt{dul:Concept} \\ \hline
  \texttt{:Situation} & \texttt{owl:equivalentClass} & \texttt{dul:Situation} \\ \hline
\end{tabular} 
\end{table}

\begin{table}[h!]
\centering
\caption{Align\-ments between the properties of SQuAP-Ont and DOLCE UltraLight.} 
\label{table:alignments_props} 
\begin{tabular}{l | c | >{\raggedleft\arraybackslash}p{4cm} }
    \textbf{SQuAP prop.} & \textbf{Align. axiom} & \textbf{DOLCE prop.}\\ \hline
    
\hline
  \texttt{:classifies} & \texttt{owl:equivalentProperty} & \texttt{dul:classifies} \\ \hline
  \texttt{:isClassifiedBy} & \texttt{owl:equivalentProperty} & \texttt{dul:isClassifiedBy} \\ \hline
  \texttt{:usesConcept} & \texttt{owl:equivalentProperty} & \texttt{dul:usesConcept} \\ \hline
  \texttt{:isConceptUsedIn} & \texttt{owl:equivalentProperty} & \texttt{dul:isConceptUsedIn} \\ \hline
  \texttt{:satisfies} & \texttt{owl:equivalentProperty} & \texttt{dul:satisfies} \\ \hline
  \texttt{:isSatisfied} & \texttt{owl:equivalentProperty} & \texttt{dul:isSatisfied} \\ \hline
  \texttt{:specializes} & \texttt{owl:equivalentProperty} & \texttt{dul:specializes} \\ \hline
  \texttt{:isSpecializedBy} & \texttt{owl:equivalentProperty} & \texttt{dul:isSpecializedBy} \\ \hline
  \texttt{:isSettingFor} & \texttt{owl:equivalentProperty} & \texttt{dul:isSettingFor} \\ \hline
  \texttt{:value} & \texttt{owl:subPropertyOf} & \texttt{dul:hasRegionDataValue} \\ \hline
\end{tabular} 
\end{table}

\subsection{Formalisation}
The following is the formalisation of SQuAP-Ont described in Section~\ref{sec:ont_description}. The formalisation is expressed in Description 
Logics. For brevity, we use the terms \texttt{Sw\-Qual\-ity\-Char} for
\texttt{Sowftware\-\-Qual\-ity\-Char\-ac\-ter\-istic}, \texttt{Arch\-Align} for 
\texttt{Ar\-chi\-tec\-tu\-ral\-Align\-ment}, \texttt{Proc\-Mat} for \texttt{Process\-Maturity},
\texttt{SwQuality} for \texttt{Soft\-ware\-Quality}, \texttt{Sw\-Qual\-ity\-Factor} for \texttt{Soft\-ware\-Qual\-ity\-Factor}, \texttt{Measure\-Res} for \texttt{Meas\-ure\-ment\-Result}, \texttt{Proc\-Mat\-Res} for \texttt{Process\-Ma\-tur\-ity\-Result}, \texttt{Sw\-Qual\-ity\-Res} for \texttt{Software\-Qual\-ity\-Result}, and \texttt{Mea\-sure\-Qual\-ity\-Res} for 
\texttt{Meas\-ure\-ment\-Quality\-Result}.

{\scriptsize
\begin{flalign}
& \mt{Value} \sqsubseteq \mt{Region} & \\
& \mt{Value} \sqsubseteq =\! 1 \mt{value}.\mt{Literal} & \\
& \mt{Concept} \not\equiv \mt{Description} & \\
& \mt{Concept} \not\equiv \mt{Description} & \\
& \mt{SwQualityChar} \sqsubseteq \mt{Concept} & \\
& \mt{SwQualityChar} \equiv \mt{ArchAlign} \sqcup \mt{ProcMat} \sqcup \mt{SwQuality} & \\
& \mt{ArchAlign} \sqsubseteq \mt{SwQualityChar}& \\
& \mt{ArchAlign} \not\equiv \mt{ProcMat} & \\ 
& \mt{ArchAlign} \not\equiv \mt{SwQuality} & \\ 
& \mt{ProcMat} \sqsubseteq \mt{SwQualityChar}& \\
& \mt{ProcMat} \not\equiv \mt{ArchAlign} & \\
& \mt{ProcMat} \not\equiv \mt{SwQuality} & \\ 
& \mt{SwQuality} \sqsubseteq \mt{SwQualityChar}& \\
& \mt{SwQuality} \not\equiv \mt{ArchAlign} & \\ 
& \mt{SwQuality} \not\equiv \mt{ProcMat} & \\ 
& \mt{Description} \not\equiv \mt{Concept} & \\
& \mt{Description} \not\equiv \mt{Situation} & \\
& \mt{SwQualityFactor} \sqsubseteq \mt{Description} & \\
& \mt{SwQualityFactor} \sqsubseteq \forall \mt{usesQualChar.SwQualityChar} & \\
& \mt{SwQualityFactor} \sqsubseteq \exists \mt{usesQualChar.SwQualityChar} & \\
& \mt{MeasureRes} \sqsubseteq \exists \mt{assess.SwQualityChar} & \\
& \mt{MeasureRes} \sqsubseteq =\! 1 \mt{hasValue}.\mt{Value} & \\ 
& \mt{MeasureRes} \sqsubseteq =\! 1 \mt{hasMetric}.\mt{Metric} & \\ 
& \mt{ArcAlignmentRes} \sqsubseteq \mt{MeasureRes} & \\
& \mt{ProcMatRes} \sqsubseteq \mt{MeasureRes} & \\
& \mt{SwQualityRes} \sqsubseteq \mt{MeasureRes} & \\
& \mt{FactorOccurrence} \sqsubseteq \mt{Situation} & \\
& \mt{FactorOccurrence} \sqsubseteq \exists \mt{isAffectedBy.MeasureRes} & \\
& \mt{FactorOccurrence} \sqsubseteq \exists \mt{satisfiesFactor.SwQualityFactor} & \\
& \mt{usesConcept} \circ \mt{specializes} \sqsubseteq \mt{usesConcept} &
\end{flalign}
}

\label{sec:ont_formalisation}
\label{sec:ont_description}

\subsection{Implementation details}
\label{sec:ont_implementation_details}
The namespace \texttt{\url{https://w3id.org/squap/}} identifies the ontology and enables permanent identifiers to be used for referring to concepts and properties of the ontology. We define individuals' URIs with the name of their types (e.g., \texttt{ArchitecturalAlignment}) preceding their IDs (e.g., \texttt{ObjectiveCharacteristic}). This convention is a common practice in many linked open data projects to define individuals' URIs. 
For example, \texttt{squap\-:\-Ar\-chi\-tec\-tural\-Align\-ment\-Ob\-ject\-ive\-Char\-ac\-ter\-istic} is the URI associated with the individual \texttt{Ob\-ject\-ive\-Char\-ac\-ter\-istic} typed as \texttt{Ar\-chi\-tec\-tural\-Align\-ment}. All the ontology entities modelled by using OWL punning follow such a convention as the can be interpreted as individuals (or classes) depending on the context. 
We setup a content negotiation mechanism that allows a client to request the ontology either (i) as HTML (e.g. when accessing the ontology via a browser) or (ii) as one of the possible serialisations allowed (i.e., RDF/XML, Turtle, N-triples). 

The alignments with DOLCE+DnS UltraLight (DUL) are published in a separate OWL file\footnote{\url{https://w3id.org/squap/squap-dul.owl}}, which imports both SQuAP-Ont and DUL. This allows one to use either SQuAP-Ont alone or its version aligned with and dependent on DUL. 
The resource, including the core ontology, the alignments, and the usage examples, is under version control on the CERN Zenodo repository\footnote{\url{http://doi.org/10.5281/zenodo.3361387}}.
SQuAP-Ont is published according to the Creative Commons Attribution 4.0 International (CC-BY-4.0) license\footnote{\url{https://creativecommons.org/licenses/by/4.0/}} and it has been uploaded on Linked Open Vocabularies\footnote{\url{https://lov.linkeddata.es/dataset/lov/}} (LOV). The license information is included in the ontology by using the \texttt{dcterms:license} property.

\section{How to use SQuAP-Ont}
\label{sec:example}
Flexibility is among the most relevant characteristic of this ontology. Although the higher levels of this ontology, regarding the standard and the factors mapping, are fixed, its measurement model can be adapted to the most suited scenario. In particular, the proposed way to evaluate information systems characteristics is just an illustrative example, which can be adapted to for any assessment purposes.

As a usage example of the SQuAP ontology, we show a real-world example consisting of the evaluation of a banking application employing the Goal-Question-Metric (GQM) approach \cite{basili1992software}. GQM defines a measurement model on three levels, i.e., Conceptual level (Goal), Operational level (Question), Quantitative level (Metric). This method offers a hierarchical assessment framework, where goals are typically defined and stable in time, and metrics may be adapted according to new measurement advances. So, we stress the fact that this paper does not focus on the measurement model, preferably on the knowledge representation of SQuAP for assessment and benchmarking purposes.
So, we provide the following synthetic RDF data about the assessment. The data are expressed as RDF serialised in TURTLE\footnote{The RDF is available at \url{https://w3id.org/squap/examples/gqm}}. 

{
\scriptsize
\begin{verbatim}
@prefix : <https://w3id.org/squap/examples/gqm/> .
@prefix arc: 
  <https://w3id.org/squap/ArchitecturalAlignment/> .
@prefix sw: 
  <https://w3id.org/squap/SoftwareQuality/> .
@prefix prc: 
  <https://w3id.org/squap/ProcessMaturity/> .
@prefix squap: <https://w3id.org/squap/> .

:compatibility-result a 
  squap:SoftwareQualityResult ;
  squap:assesses sw:Compatibility ;
  squap:hasMetric :sonarqube-sw-quality ;
  squap:hasValue :sonarqube-value-b .
  
:correspondenceresult 
  a squap:ArchitecturalAlignmentResult ;
  squap:assesses arc:Correspondence ;
  squap:hasMetric :likert-scale-1-7 ;
  squap:hasValue :likert-value-7 .
  
:documentation-result 
  a squap:ProcessMaturityResult ;
  squap:assesses prc:Documentation ;
  squap:hasMetric :likert-based-prc-maturity ;
  squap:hasValue :likert-value-6 .


:sonarqube-sw-quality a squap:Metric ;
  squap:hasParameter :sonarqube-params .
  
:sonarqube-params a squap:Parameter ;
  squap:parametrizes :sonarqube-value-a , 
    :sonarqube-value-b , 
    :sonarqube-value-c .

:likert-based-prc-maturity a squap:Metric ;
  squap:hasParameter :likert-scale-1-7 .
  
:likert-scale-1-7 a squap:Parameter ;
  squap:parametrizes 
    :likert-value-1 , :likert-value-2 , 
    :likert-value-3 , :likert-value-4 , 
    :likert-value-5 , :likert-value-6 , 
    :likert-value-7 . 

:sonarqube-value-b a squap:Value ;
  squap:value "B" .

:likert-value-7 a squap:Value ;
  squap:value 7 .
  
:likert-value-6 a squap:Value ; 
  squap:value 6 .
\end{verbatim}
}

\begin{figure}[ht!] \centering
\includegraphics[scale=0.44]{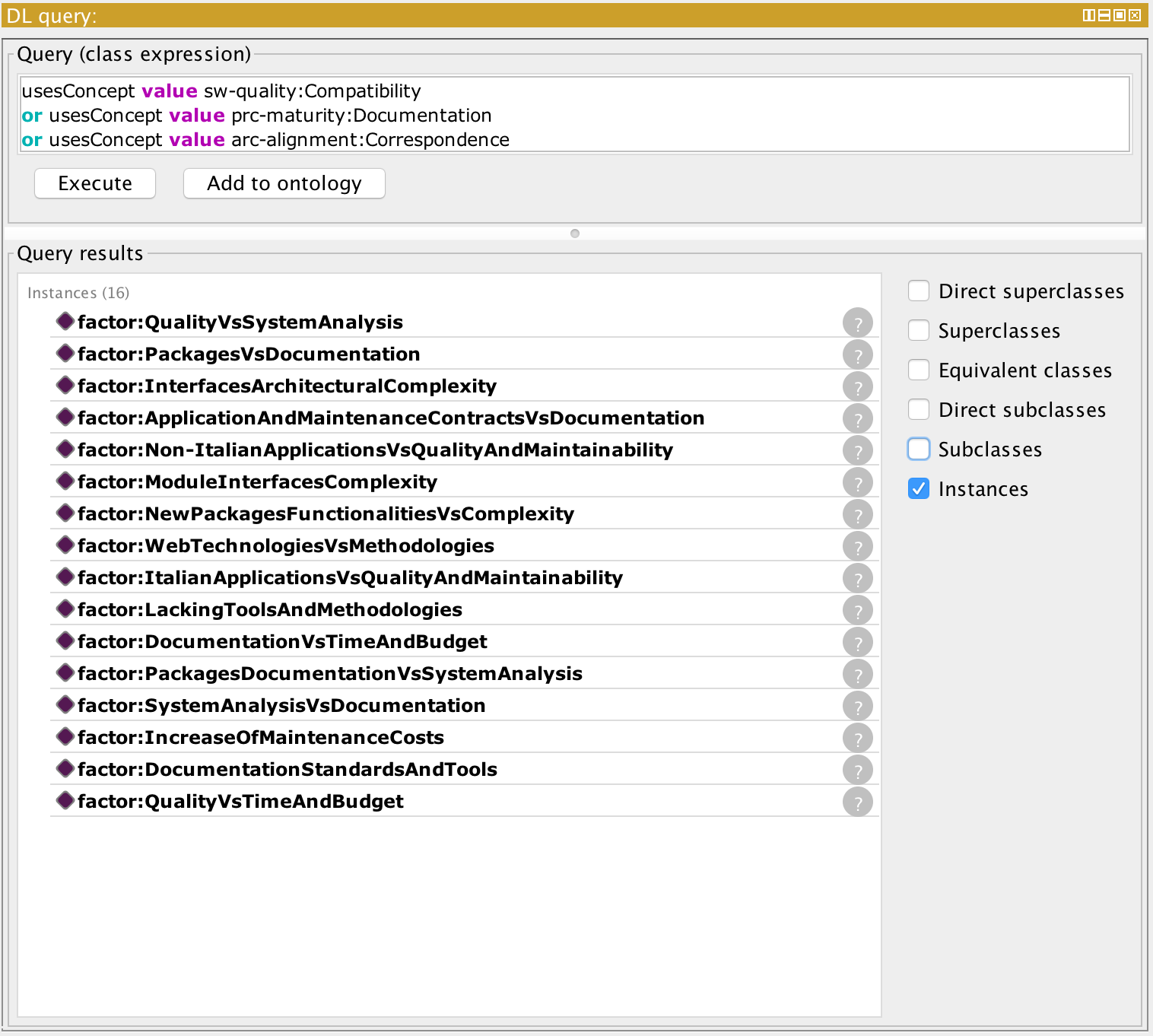}
    \caption{Execution of a DL query on the RDF sample.}
    \label{fig:dl-query}
\end{figure}

The example describes a banking system associated with three assessments about the dimensions of software quality, architectural alignment, and process maturity. The specific measurement results  are: \texttt{:compatibility-result}, which assesses the characteristic \texttt{sw-qua\-li\-ty:\-Com\-pat\-ib\-il\-ity} (software quality), \texttt{:cor\-res\-pond\-ence\-res\-ult}, which assesses the characteristic \texttt{arc-align\-ment:Cor\-res\-pond\-ence} (architectural alignment), and \texttt{:doc\-u\-ment\-a\-tion\--res\-ult}, which assesses the characteristic \texttt{prc-ma\-tur\-ity:\-Doc\-u\-ment\-a\-tion} (process maturity). Those measurement results are associated with a value (e.g., \texttt{:likert-value-7}, which identifies the value 7 of a Likert scale) and a metric (e.g., \texttt{:likert-scale-1-7}, which identifies a Likert scale ranging from 1 to 7). Each value is reported with a literal representation and is associated with a metric. It is possible to use the axioms defined in SQuAP-Ont in order to gather all the factors that can be enabled by the available measured quality characteristics (e.g., \texttt{sw-qua\-li\-ty:\-Com\-pat\-ib\-il\-ity}). This can be done, for example, by executing a Prot\'{e}g\'{e} DL query, the result of which is shown in Figure~\ref{fig:dl-query}.

In this example, different standards' items represent the Goals, which are measured with one or several (also concurrent) software quality metrics.
To do so, we followed literature recommendations \cite{Wagner2012,campbell2013sonarqube}.
The result is the sum of different evaluators, which represent a measurement of the three standards. 


Alternatively, it is possible to define productive rules to materialise the factors that are enabled by the available measured quality characteristics. The following SPARQL CONSTRUCT is a possible productive rule for our example.

{
\scriptsize
\begin{verbatim}
PREFIX squap: <https://w3id.org/squap/>
PREFIX rdfs: <http://www.w3.org/2000/01/rdf-schema#>
CONSTRUCT {
  ?measurementResult 
    squap:affectsMeasurementOf ?factorOccurrence . 
  ?factorOccurrence 
    a squap:FactorOccurrence;
    squap:satisfiesFactor ?factor
}
WHERE{
  ?factor 
    squap::usesQualityCharacteristic ?char;
    rdfs:label ?factorLabel . 
  ?measurementResult 
    squap:assesses ?char
  BIND(IRI(
    CONCAT("https://w3id.org/squap/example/gqm/", 
      ?factorLabel)) 
    AS ?factorOccurrence)
}
\end{verbatim}
}

We remark that factors and quality characteristics are defined in SQuAP-Ont both as classes and individuals through OWL punning. Hence, one can decide to use DL reasoning or rules defined in any other formalism depending by the specific case, e.g., SPARQL CONSTRUCT, Shapes Constraint Language\footnote{\url{https://www.w3.org/TR/shacl/}} (SHACL), etc. 

Another worth presenting example is a {\em dogfooding} usage scenario. Dogfooding is when an organization uses its product for demonstrating its quality. In this example, we use the SQuAP-Ont to record the metrics, the values, the factors, and the quality characteristics resulting from the measurement of its characteristics. The data of this example are expressed as RDF serialised in TURTLE.

{
\scriptsize
\begin{verbatim}
@prefix : 
  <https://w3id.org/squap/examples/dogfooding/> .
@prefix owl:
  <http://www.w3.org/2002/07/owl#> .
@prefix rdfs:
  <http://www.w3.org/2000/01/rdf-schema#> .
@prefix xsd:
  <http://www.w3.org/2001/XMLSchema#> .
@prefix squap:
  <https://w3id.org/squap/> .
@prefix factor:
  <https://w3id.org/squap/Factor/> .
@prefix prc:
  <https://w3id.org/squap/ProcessMaturity/> .

:documentation-measurement-result
  a  squap:MeasurementResult ;
  squap:hasMetric
    :protege-ontology-annotations-metric ;
  squap:hasValue
    :documentation-measurement-result-value .

:documentation-measurement-result-value
  a  squap:Value ;
  squap:value "233"^^xsd:integer .
  
:protege-ontology-annotations-metric
  a  squap:Metric ;
  squap:assess
    prc:Documentation 

prc:Documentation
  a  squap:ProcessMaturity .

factor:PackagesVsDocumentation
  a  squap:Factor ;
  squap:usesConcept prc:Documentation .

:process-maturity-occurrence
  a  squap:FactorOccurrence ;
  squap:isAffectedBy
    :documentation-measurement-result ;
  squap:satisfiesFactor
    factor:PackagesVsDocumentation .
    
\end{verbatim}
}

In the dogfooding example the ontology is used for recording a measurement result (i.e., \texttt{:do\-cu\-men\-ta\-tion-mea\-su\-re\-ment-re\-sult}) for the SQuAP-ONT based on the Prot\'{e}g\'{e} ontology metric that records the number of annotations in an ontology (i.e., \texttt{:pro\-te\-ge-on\-to\-lo\-gy-an\-no\-ta\-tions-me\-tric}). The value associated with such a measurement result is "233", that is the number of annotations used for documenting the ontology. The aforementioned value can be easily obtained by using the ontology metrics view when opening the SQuAP-ONT with Prot\'{e}g\'{e}. The \texttt{:pro\-te\-ge-on\-to\-lo\-gy-an\-no\-ta\-tions-me\-tric} assesses the \texttt{:prc:Do\-cu\-men\-ta\-tion}, which is a specific concept defined in SQuAP-ONT for characterising the process maturity (i.e., \texttt{squap:Pro\-cess\-Ma\-tu\-ri\-ty}) in terms of how much the software is documented. It is worth noticing that, by relying on OWL punning, we use \texttt{prc:Do\-cu\-men\-ta\-tion} as an individual, though it is also defined as a class in the ontology. The concept of \texttt{prc:Do\-cu\-men\-ta\-tion} is used by a specif factor, that is \texttt{fa\-ctor:Pa\-cka\-ges\-Vs\-Do\-cu\-men\-ta\-tion}. Accordingly, we have a factor occurrence, that is \texttt{:pro\-cess-ma\-tu\-ri\-ty-oc\-cur\-rence}.

\section{Potential impact}
\label{sec:impact}
In the last decade, there has been a considerable effort, especially by the Management Information Systems research community, to study the phenomenon of the alignment of business and information systems \cite{aerts2004architectures}. What emerged is the importance of such alignment for both business' competitiveness and technical efficiency. When it comes to integrating new solutions, modules, or interfaces, such alignment is of crucial importance.
Several other scholars found similar results, suggesting the importance of standard governance defining key architecture roles, involving critical stakeholders through liaison roles and direct communication, institutionalizing monitoring processes and centralizing IT critical decisions \cite{boh2006using}. Especially in the financial sector, architectural governance is a crucial issue for IT efficiency and flexibility \cite{schmidt2011outcomes}.
Generally speaking, this finding is also primarily shared beyond the financial sector \cite{lange2016empirical}. The need for people from different backgrounds (mainly business and technical ones) to align the organization is the most considerable insight into this research stream.

To tackle the issue of information systems quality from an empirical perspective, we started in 2014 to survey banking application maintenance group experts, Chief Executive Officers, Chief Information Officers, IT architects, technical sales accounts, Chief Data Officers, and maintenance managers \cite{Russo2017metaquality}.
This ongoing project is pursued with a leading consultancy firm, according to which we were able to cover with our representative sample the IT banking sector.
Consequently, the need for knowledge representation of different measurement models is perceived as contingent and requested by the IT banking community in this significant project on information systems' quality. One crucial insight that emerged from the factors is the difficulty to assess their applications, also due to the diversity and complexity of measurement models.
Indeed, the three standards measure three different dimensions.
Quality measures the software as a product; Process as a process; and Architecture the alignment to a taxonomy. Accordingly, metrics and predictors reflect these differences.
Therefore, the development of this ontology is a direct request from practitioners.

Since this research journey started from an industry's need, an ontology, intended as the knowledge representation of different measurement models is of pivotal importance, and a first tool to systematize the assessment of banking information systems' quality.
Thus, this ontology will be used for consultancy purposes to implement the SQuAP quality model.
Moreover, it is also useful to trace changes in quality in time and suggest specific improvements. So, this ontology is the knowledge layer over which this quality model is built.
Consultancy firms expressed their interest in a knowledge representation tool which can be displayed to customers in the assessment phase, to tailor their consultancy efforts. However, also the bank's IT departments will use it for similar purposes.
They can also tailor-made and modify this ontology and the underlying metrics suggested by the literature, according to their specific needs.

For this reason, we used a CC-BY-4.0 license, open to commercial use.
Our industrial partners consider the use and reuse of this ontology as an excellent value for the practitioners' community.

\section{Conclusion and future development}
\label{sec:conclusions}
In this paper, we have described SQuAP-Ont, an ontology to assess information systems of the banking sector. SQuAP-Ont  a) guides its users through the (ongoing) assessment phases suggested by software engineering literature;
b) helps to identify critical quality flaws within applications; and c) extends and integrates existing work on software ISO ontology terms, diagram visualizations and ontology revisions. 
SQuAP-Ont has been developed for commercial use, within an industrial project on quality of banking information systems. Nevertheless, like all ontologies, it is an evolving effort, and we are open to suggestions proposed by the broad researchers' and practitioners' communities.
We have addressed several issues raised in previous studies, and according to the industry's expectations.

Our future work aims to facilitate enrichment and refine the ontology continuously along with standards and literature recommendation changes. The enrichment is also about the introduction of specific annotations based on reference vocabularies for tracking provenance and versioning (e.g., PROV-O). Another important aspect is to validate and monitor the application of SQuAP in domains and software projects different from the banking system context. This may lead to the model's enrichment and improvement.
 \section*{Acknowledgments}
 This work was partially funded by the Consorzio Interuniversitario Nazionale per l'Informatica (CINI)
 and the Science Foundation Ireland grant 15/SIRG/3293 and 13/RC/2094 and co-funded under the European Regional Development Fund through the Southern \& Eastern Regional Operational Programme to 
Lero---the Irish Software Research Centre. (www.lero.ie).

\bibliographystyle{aps-nameyear}      
\bibliography{bibliography.bib}                 
\nocite{*}
\end{document}